\pdfoutput=1

\documentclass[11pt]{article}

\usepackage[]{emnlp2021}

\usepackage{times}
\usepackage{graphicx}
\usepackage{url}
\usepackage{latexsym}
\usepackage{subcaption}
\usepackage{hyperref}
\usepackage{multirow}
\usepackage[T1]{fontenc}

\usepackage[utf8]{inputenc}

\usepackage{microtype}

\title{Text Simplification for Comprehension-based Question-Answering}

\author{Tanvi Dadu \thanks{\hspace{2 mm} The first two authors contributed equally to the work.} \thanks{\hspace{2 mm} This work was done when Kartikey Pant was affiliated to IIIT Hyderabad, Tanvi Dadu was affiliated to NSIT Delhi, and Kuntal Dey was affiliated to IBM.} \\
NSIT, Delhi \\
\newline \\ 
\textbf{Ferdous Ahmed Barbhuiya}\\ 
IIIT, Guwahati \\
\\\And
Kartikey Pant \footnotemark[1] \footnotemark[2]\\
IIIT, Hyderabad \\
\newline \\ 
\textbf{Kuntal Dey \footnotemark[2]}\\ 
Accenture Tech Labs \\
\And 
\textbf{Seema Nagar}\\ 
IBM Research India \\
}

\begin{document}
\maketitle
\begin{abstract}
Text simplification is the process of splitting and rephrasing a sentence to a sequence of sentences making it easier to read and understand while preserving the content and approximating the original meaning. Text simplification has been exploited in NLP applications like machine translation, summarization, semantic role labeling, and information extraction, opening a broad avenue for its exploitation in comprehension-based question-answering downstream tasks. In this work, we investigate the effect of text simplification in the task of question-answering using a comprehension context. We release \textit{Simple-SQuAD}, a simplified version of the widely-used \textit{SQuAD} dataset. 

Firstly, we outline each step in the dataset creation pipeline, including style transfer, thresholding of sentences showing correct transfer, and offset finding for each answer. Secondly, we verify the quality of the transferred sentences through various methodologies involving both automated and human evaluation. Thirdly, we benchmark the newly created corpus and perform an ablation study for examining the effect of the simplification process in the \textit{SQuAD}-based question answering task. Our experiments show that simplification leads to up to $2.04\%$ and $1.74\%$ increase in \textit{Exact Match} and \textit{F1}, respectively. Finally, we conclude with an analysis of the transfer process, investigating the types of edits made by the model, and the effect of sentence length on the transfer model.   
\end{abstract}

\section{Introduction}
\label{intro}
Text simplification is the task of modifying the structure of a text to make it easier to read and comprehend while preserving the content and approximating the original meaning. Linguistically, simple sentences are defined as having only one subject and one verb or predicate. Therefore, a complex sentence can be rewritten into multiple simpler sentences while retaining the same meaning. When simplifying texts, a myriad of rewriting transformations have been explored, ranging from replacing complex words or phrases for simpler synonyms, to changing the syntactic structure of the sentence. Moreover, modern automated text simplification approaches are data-driven, attempting to simplify sentences using parallel corpora of aligned complex-simplified sentences.

Text simplification allows humans to read texts more efficiently and faster, empowering them to take part in society and become more active in their daily actions and healthcare \cite{Dannells}. Another vital application of text simplification is reading assistance it provides, especially for people with reading disabilities \cite{Carroll1999,Inui2003}, low-literacy readers \cite{Watanabe2009}, or non-native speakers \cite{Siddhartha2002}.

\begin{table*}[h]
\centering
\resizebox{0.85\textwidth}{!}{%
\begin{tabular}{|l|l|}
\hline
\textbf{Original SQuAD Sentence} &
  \textbf{Transferred Simple-SQuAD Sentence} \\ \hline
\begin{tabular}[c]{@{}l@{}}Clark also claimed that Abdul gave him preferential \\ treatment on the show \underline{due} to their affair.\end{tabular} &
  \begin{tabular}[c]{@{}l@{}}Clark also claimed that Abdul gave him preferential\\ treatment on the show\underline{. This was due} to their affair.\end{tabular} \\ \hline
\begin{tabular}[c]{@{}l@{}}In his Prices and Production (1931), Hayek argued \\ that the business cycle resulted from the central bank's\\ inflationary credit expansion and its transmission over \\ time\underline{, leading} to a capital misallocation caused by the \\ artificially low interest rates.\end{tabular} &
  \begin{tabular}[c]{@{}l@{}}In his Prices and Production (1931), Hayek argued\\ that the business cycle resulted from the central bank's\\ inflationary credit expansion and its transmission over \\ time\underline{. This led} to a capital misallocation caused by the \\ artificially low interest rates.\end{tabular} \\ \hline
\end{tabular}
}
\caption{Examples of Simple and Complex sentences, from our proposed Simple-SQuAD dataset and the SQuAD dataset. Underlined parts denotes the splitting point.}
\label{tab:simple-introduction}
\end{table*}

There has been considerable work done where NLP applications exploit the benefits of text simplification. Long sentences with complex syntax or those laden with long-distance dependencies often pose difficulties for various downstream tasks. Text simplification has been used to improve the performance for these tasks including machine translation \cite{Hasler2017}, summarization \cite{Silveira2012}, semantic role labeling \cite{Vickrey2008,Evans2019}, and information extraction \cite{Evans2019}. This opens up the avenue for exploring simplification for question-answering. To the best of our knowledge, text simplification for comprehension-based question-answering has not been explored yet.

In this work, we make the following contributions:-
\begin{itemize}
\item We propose a transformers-based text-simplification pipeline that splits and rephrases a complex sentence into its simpler constituent sentences. We outline each step in the dataset creation pipeline, including data preprocessing, performing text simplification, thresholding the simplified sentence based on the quality of the transfer, and offset finding of answers from each question-answer pair present in the dataset.
\item We propose a new dataset \textit{Simple-SQuAD} by converting each context in the \textit{SQuAD} dataset using the proposed text-simplification pipeline.  \footnote{Made available at the following Github repository: \url{https://github.com/kartikeypant/text-simplification-qa-www2021}.}
\item We perform automated and human evaluation to determine the quality of the text simplification model.
\item We perform event based analysis and sentence-length based transfer analysis to give deeper insights into the transfer process.
\item We then benchmark both \textit{Simple-SQuAD} and \textit{SQuAD} for predictive performance in the \textit{Simple-SQuAD} question-answering task.
\end{itemize}

\section{Related Works}

\subsection{Text Simplification}

Text simplification has attracted a great deal of attention due to its potential impact on society. Prior explorations on text simplification contain a myriad of approaches, which include from the use of hand-crafted syntactic rules \cite{Carroll1999,Vickrey2008}, statistical simplification model \cite {Zhu2010}, quasi-synchronous grammar \cite{Woodsend2011} and the semantic hierarchy of simplified sentences for recursively splitting and rephrasing complex sentences \cite{Niklaus2019}. Recently, Machine Translation, both statistical and neural, has also been used for the task of text simplification \cite{Narayan2014,Nisioi2017}.

The process of text simplification has been observed to help in the performance of multiple downstream tasks. Silveira et al.~\cite{Silveira2012} explored the use of a sentence simplification module in summarization systems, concluding that the simplification module removes expendable information that helps in accommodating relevant data in a summary. Hasler et al.~\cite{Hasler2017} showed that the performance of source simplification could improve translation quality in machine translation systems. Evans et al.~\cite{Evans2019} proved the efficacy of integrating a text simplification step for improving the predictive performance of semantic role labeling and information extraction methodologies.

Narayan et al.~\cite{Narayan2017} introduced a new text simplification task, named Split-and-Rephrase on their proposed WebSplit dataset, containing $1,066,115$ parallel instances of complex and sequence of simple sentences having a similar meaning. The goal of the task is to split a complex input sentence into shorter sentences while preserving the meaning. In this task, the emphasis is on sentence splitting and rephrasing, with no deletion and no lexical or phrasal simplification. They further proposed five models, ranging from vanilla sequence-to-sequence to semantically-motivated models to benchmark the proposed task. Aharoni and Goldberg~\cite{Aharoni2018} and Botha et al.~\cite{Botha2018} extended the work by introducing new datasets, with a more extensive vocabulary and split examples to improve the efficacy of the prior benchmarks. We use WikiSplit, introduced by Botha et al.~\cite{Botha2018}, for training our sentence simplification module.

\subsection{Style Transfer}
In recent works, textual style transfer has been shown to produce grammatically fluent, and information-preserved texts with fairly accurate target attributes. For the task in a semi-supervised setting, various methodologies are exploited, including back-translation~\cite{Prabhumoye2018}, back-translation with attribute conditioning~\cite{Pant2020}, specialized transfer methodologies like Delete, Retrieve, Generate~\cite{Li2018}. However, in the presence of a large parallel corpus, sequence-to-sequence models perform competitively. Aharoni and Goldberg~\cite{Aharoni2018} exploited a copy-mechanism based sequence-to-sequence model with attention~\cite{Bahdanau2015} for the text simplification transfer.

Transformers~\cite{Vaswani2017} have been shown to perform robust language modeling, given enough data, helping in various downstream tasks.  Due to the parallelized nature of the architecture, it is possible to train using much larger datasets than recurrent neural networks. However, it becomes necessary to optimize hyperparameters carefully while ensuring scalability ~\cite{Popel2018}. Transformer-based methodologies for the task of sentence simplification have been explored by Marayuma and Yamamoto~\cite{Maruyama2019} for Japanese and Zhao et al.~\cite{Zhao2018} for a smaller Wikipedia-based dataset. However, to the best of our knowledge, there has been no work done for supervised style transfer exploiting transformer-based models for the significantly large \textit{WikiSplit} dataset.

\section{Corpus Creation}

\subsection{Preliminaries}

\subsubsection{\textbf{SQuAD}}

For exploring the effect of text simplification in the question-answering downstream task, we use Stanford Question Answering Dataset (SQuAD comprehension), a reading comprehension dataset released by Rajpurkar et al. ~\cite{Rajpurkar2016}. It consists of $100,000+$ questions from $536$ articles posed by crowdworkers on a set of Wikipedia articles, where the answer to each question is a segment of text from the corresponding reading passage.  Unlike other datasets, \textit{SQuAD} does not provide a list of answer choices for each question; instead, systems must select the answer from all possible spans in the passage. Though the system must deal with a large number of candidate answers, yet span-based answers are easier to evaluate than free-form answers. Further, it is diverse in terms of answer types, containing a significant percentage of dates, numeric data, adjective phrases, verb phrases, clauses as answers. The predictive performance of models in the \textit{SQuAD} dataset is evaluated using \textit{Exact-match} and \textit{F1-score}.

\subsubsection{\textbf{WikiSplit}}

For the task of text simplification, we use the \textit{WikiSplit} corpus \cite{Botha2018}, a parallel corpus consisting of complex sentences and its consequent sequence of simple sentences having similar meaning for the Split and Rephrase task. It contains a set of one million naturally occurring sentence rewrites mined from English Wikipedia, providing $60$ times more examples and $90$ times more vocabulary as compared to the \textit{WebSplit} corpus introduced by Narayan et al. ~\cite{Narayan2017}. Using a larger dataset for the task of text simplification increases the efficacy of the model and the quality of the transferred data ~\cite{Botha2018,Popel2018}.

\begin{figure*}[h]
    \centering
    \includegraphics[width=1\textwidth]{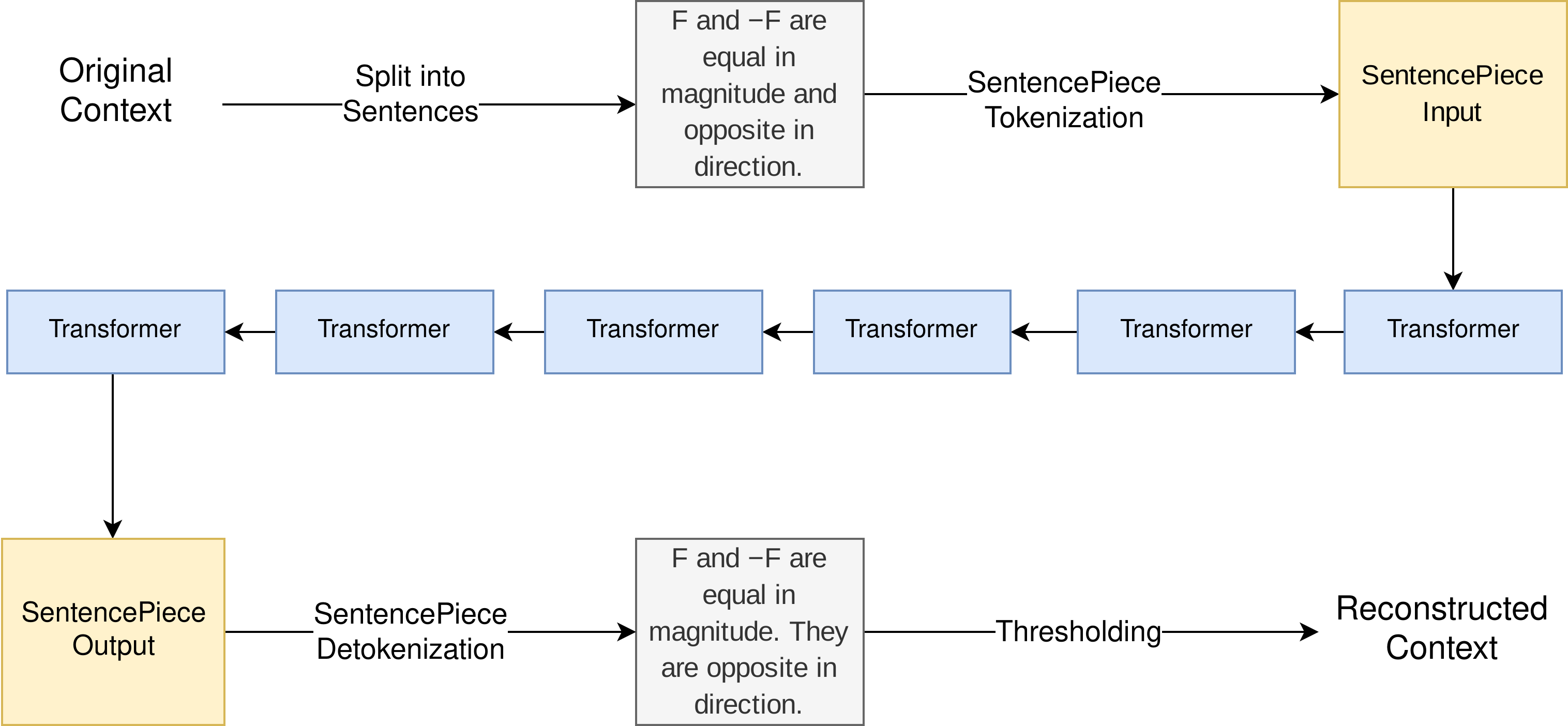}
    \caption{Style Transfer Architecture}
    \label{fig:architecture-diagram}
\end{figure*}

\subsection{Approach}
\subsubsection{\textbf{Style Transfer}}

This subsection outlines the process of textual style transfer using transformers for converting the complex contexts in the \textit{SQuAD} dataset into their simpler counterparts. Figure~\ref{fig:architecture-diagram} illustrates the process for a sample context and question-answer pair from the \textit{SQuAD} dataset.

We use spaCy’s Sentencizer \footnote{\href{https://spacy.io/api/sentencizer}{https://spacy.io/api/sentencizer}} to tokenize the contexts into their respective constituent sentences.  This process enables us to perform the sentence-level transfer, thus ensuring higher degrees of overall transfer quality. We tokenize each sentence using a SentencePiece tokenizer, trained on $3.24$ \textit{GBs} English Wikipedia . While training the SentencePiece tokenizer, we mark the numerical tokens as custom-defined ones, ensuring no tokenization happens for such tokens. This step helps in preserving the numerical tokens in the transfer process by the transfer model.

We transfer each tokenized sentence using a transformers based machine translation model. We use the OpenNMT-py \cite{OpenNMT_Paper} toolkit for the implementation process. The model consists of $6$-layered transformers architecture with $8$ self-attention heads and $1028$ sized hidden feed-forward layer. We trained the model for $20000$ training steps with a dropout of $0.1$ and a batch size of $2048$. For optimization, we used Adam optimizer with a \textit{beta2} value of $0.998$, accumulating the gradient twice. We use an initial learning rate of $2$ with \textit{noam} decay method and $800$ ($4\%$ of total) warmup steps.

\subsubsection{\textbf{Thresholding}}
In this section, we outline the thresholding techniques used in the generated sequence of simple sentences to maintain their quality.  We execute the following post-editing steps to reduce noise from the dataset:
\begin{enumerate}
\item \textbf{Perplexity} : We use open \textit{GPT-2} \cite{Radford2018} as the language model to assign perplexity to the generated sentences. We threshold the transferred sequence of sentences that have the perplexity between $50$ and $600$ and discard all the remaining sequences. This thresholding ensures that the sequence of sentences is fluent, as measured by the language model. Table \ref{tab:thresholding-split} shows that $84951$ of $91757$ sentences are within the fluency thresholding, comprising of $~93\%$ of the total sentences.
\item \textbf{Length of the original sentence}: We weed out all sentences whose original length was shorter than five tokens. This heuristic is based on the condition that sentences lesser than five tokens are highly likely to be simple, reducing the number of false positives to a significant extent efficiently. Table \ref{tab:thresholding-split} shows that the thresholding process removes $15.16\%$ of the already thresholded sentences leaving $~78.5\%$ of the total sentences.
\item \textbf{Redundancy in sequence of the transferred sentences}: We remove all transferred sentences that display a redundancy behavior, which is defined to be one in which two or more sentences in the sequence are the same. We introduce this step after a careful analysis of the transferred sentences. Table \ref{tab:thresholding-split} illustrates that the thresholding removes $1.57\%$ of the already thresholded sentences leaving $~77.31\%$ of the total sentences.
\end{enumerate}
\begin{table}[]
\centering
\resizebox{0.48\textwidth}{!}{%
\begin{tabular}{|l|r|r|}
\hline
\begin{tabular}[c]{@{}l@{}}\textbf{Thresholding} \\ \textbf{Step}\end{tabular} &
  \begin{tabular}[c]{@{}r@{}}\textbf{Percentage of} \\ \textbf{Total Sentences}\end{tabular} &
  \begin{tabular}[c]{@{}r@{}}\textbf{Thresholded} \\ \textbf{Sentences}\end{tabular} \\ \hline
\textbf{Step 1}: Perplexity                                                              & 92.58\% & 84,951 \\ \hline
\begin{tabular}[c]{@{}l@{}}\textbf{Step 2}: Original \\ Length Thresholding\end{tabular} & 78.5\%  & 72,075 \\ \hline
\begin{tabular}[c]{@{}l@{}}\textbf{Step 3}: Redundancy \\ Thresholding\end{tabular}      & 77.31\% & 70,944 \\ \hline
\end{tabular}%
}
\caption{Sentences left after each thresholding process step}
\label{tab:thresholding-split}
\end{table}

\subsubsection{\textbf{Automated Evaluation}}
We perform an automated sentence-level analysis of the style transfer model. We determine the extent of content preservation, lexical simplicity, and reading ease of the generated sentences using the following three metrics:

\begin{enumerate}
    \item \textbf{BLEU} (Papineni et al.~\cite{Papineni2002}): We use the self-BLEU score taking the original sentence as the reference to measure the extent of content preservation. Table \ref{tab:metric-distribution} shows that our model achieves a mean high self-BLEU of $75.66$, implying that the transferred sentences display a high level of content-based similarity with the original sentence
    \item \textbf{SARI} (Xu et al.~\cite{Xu2016}): We use the SARI metric to measure the quality of lexical simplicity in the transferred sentences. It analyzes the words added, deleted, and retained by a simplification model. Our version of SARI compares the model’s output to the original sentence. There is a high correlation with human judgments of simplicity gain and the SARI metric (Xu et al., 2016). In Table \ref{tab:metric-distribution}, we observe a mean SARI value of $30.08$ with a low standard deviation of $2.21$, denoting most sentences to have a high level of lexical simplicity.
    \item \textbf{Flesch–Kincaid Grade level} (Kincaid et al.~\cite{Kincaid1975}): Flesch– Kincaid Grade Level (FKGL) is a widely-used metric for text readability. It represents the corresponding U.S. grade level, whose education is appropriate for understanding the text. As illustrated in Table \ref{tab:metric-distribution}, the Flesch-Kincaid Grade Level for the sentences decreases by $5.43$ grade levels on average. Thus, the transferred sentences have much higher readability than the original sentences.

\end{enumerate}

\begin{table}[h]
\centering
\resizebox{0.46\textwidth}{!}{%
\begin{tabular}{|l|r|r|}
\hline
\textbf{Metric} & \textbf{Mean} & \textbf{Standard Deviation} \\ \hline
\textbf{BLEU} & 75.66 & 13.68 \\ \hline
\textbf{SARI} & 30.08 & 2.21 \\ \hline
\textbf{FKGL(Original)} & 13.29 & 5.74 \\ \hline
\textbf{FKGL(Transferred)} & 7.86 & 3.94 \\ \hline
\end{tabular}%
}
\caption{Automated analysis of transferred sentences}
\label{tab:metric-distribution}
\end{table}

\subsubsection{\textbf{Human Evaluation}}

Although we use commonly used metrics for the evaluation, automated evaluation of generative models of text is still an open research problem ~\cite{Hu2017}. We perform a human evaluation to analyze the quality of the transferred sentences accurately. 

In the human evaluation, we randomly sampled $50$ sentence pairs from the thresholded dataset consisting of both transferred sentences and their original variants. The human evaluators were asked to rate each sentence pair on a $1$-$5$ Likert scale on the following metrics: 

\begin{enumerate}
    \item \textbf{Fluency}: A high \textit{Fluency} score, $4$ or $5$, denotes that the transferred sentence is well constructed. A medium score of $3$ denotes that the sentence contains lexical errors, while a low score of $2$ or $1$ denotes major errors and extremely poor constructions respectively.
    \item \textbf{Relative Simplicity}: A sentence pair is given a high \textit{Relative Simplicity} score, $4$ or $5$, if the transferred sentence is significantly simplified as compared to the original sentence. If the simplicity remains the same, it is given a medium score of $3$ score. Moreover, a low score represents that the transferred sentence is more complex than the original sentence.
    \item \textbf{Content Preservation}: A high \textit{Content Preservation} score of $4$ and $5$ indicates that the content of the original sentence was well-preserved in the transferred sentence. A medium score of $3$ denotes that there were minor differences between the transferred sentences and the original sentences. A low score of $2$ and $1$ denotes major violation of content preservation. 
\end{enumerate}

To verify the inter-rater agreement, we perform a Krippendorff's alpha analysis across all three metrics. Our analysis shows that we obtain an averaged Krippendorff's alpha inter-rater agreement of $0.63$ over all metrics, denoting reasonable agreement between the raters.
\begin{table}[h]
\centering
\begin{tabular}{|l|r|}
\hline
\textbf{Metric}               & \textbf{Average Value} \\ \hline
\textbf{Fluency}              & 4.26                   \\ \hline
\textbf{Relative Simplicity}  & 3.91                   \\ \hline
\textbf{Content Preservation} & 4.10                   \\ \hline
\end{tabular}
\caption{Human Evaluation Results}
\label{tab:human-evaluation}
\end{table}

Table \ref{tab:human-evaluation} illustrates the results of the performed human evaluation. We observe that the transferred sentences were judged to be fluent, significantly simplified, and content-preserved compared to their original counterparts.

\subsubsection{\textbf{Offset Finding and Dataset Finalization}}
In this subsection, we outline the process of finding the offsets for each question-answer pair and reconstructing our proposed simplified version of \textit{SQuAD}, \textit{Simple-SQuAD}, from the thresholded sentences.

Firstly, we preprocess the sentences, which involves removing the split sentence indicator and reconstructing context texts from both thresholded and original sentences. If a sentence is not simplified, we use the original sentence itself, ensuring minimal loss of information due to the transfer process. 

Secondly, for every question-answer pair, we calculate the answer’s character offset by using exact matching in the reconstructed context. If we do not find an exact match, we use case-insensitive pattern matching to calculate the offset. We found the offsets for $88,690$ questions in total. 

Finally, we create the following two datasets: \textit{Simple-SQuAD} and \textit{Original}. The \textit{Simple-SQuAD} dataset contains the reconstructed context with the calculated character offsets for each answer. The \textit{Original} dataset contains the original context itself but with only those questions whose answers are present in the \textit{Simple-SQuAD} dataset. We create the new \textit{Original} dataset to ensure an equal number of question-answer pairs, leading to a fair comparison. This \textit{Original} dataset represents the \textit{SQuAD} candidate for the benchmarking process.

\section{Benchmarking Experiments}

This section highlights the models and experiments performed for benchmarking the \textit{Simple-SQuAD}. We compare the results obtained for \textit{Simple-SQuAD} against the \textit{SQuAD} dataset. 

\subsection{Model Used}

For benchmarking \textit{Simple-SQuAD}, we use two different variations of RoBERTa as introduced by Liu et al.~\cite{Liu2019RoBERTaAR}. RoBERTa is a replication study of BERT pretraining, which is trained on more extensive training data with bigger batches, longer sequences, and dynamically changing masking patterns. Consequently, RoBERTa achieves better results over BERT and attains state-of-the-art results on \textit{GLUE}, \textit{RACE}, and \textit{SQuAD}.

\subsection{Experimental Settings}

We perform a dataset-based ablation study, experimenting with multiple variants of input datasets for each model. Firstly, we finetune the model on the \textit{SQuAD} and the \textit{Simple-SQuAD} dataset separately for $2$ epochs. We then finetune the \textit{Simple-SQuAD} trained model on the \textit{SQuAD} dataset and the \textit{SQuAD}-trained model on the \textit{Simple-SQuAD} dataset for $2$ epochs each. We benchmark the results for each of these combinations of the dataset input to better infer the effect of simplifying sentences in the original dataset.

For benchmarking, we use $442$ training articles containing $78,810$ questions and $48$ development articles containing $9880$ questions. Thus, we have a $90$:$10$ and an approximate $89$:$11$ train-test split based on the number of articles and the number of questions, respectively. We used $10\%$ of the training examples as a validation set for both our models.

For both $RoBERTa_{Base}$ and $RoBERTa_{Large}$, we use a maximum sequence length of $380$, a stride of $128$, a maximum query length of $64$, and a maximum answer length of 30. We use a learning rate of $1*10^{-5}$ with a weight decay of $0.01$.

\subsection{Results}
This section outlines the result of the ablation study to determine the effect of text simplification on the question answering downstream task in the \textit{SQuAD} dataset. We observe that text simplification improves the predictive performance of both $RoBERTa_{base}$ and $RoBERTa_{large}$.  

\begin{table}[h]
\centering
\resizebox{0.47\textwidth}{!}{%
\begin{tabular}{|l|l|r|r|}
\hline
\textbf{Model} & \textbf{Input} & \textbf{Exact} & \textbf{F1} \\ \hline
$RoBERTa_{Base}$ & SQuAD & 0.787 & 0.863 \\ \hline
$RoBERTa_{Base}$ & Simple-SQuAD & 0.786 & 0.866 \\ \hline
$RoBERTa_{Base}$ & Simple-SQUAD $\to$ SQuAD & 0.799 & 0.876 \\ \hline
$RoBERTa_{Base}$ & SQuAD $\to$ Simple-SQuAD & \textbf{0.803} & \textbf{0.878} \\ \hline
\end{tabular}%
}
\caption{Benchmarking Results for $RoBERTa_{Base}$.}
\label{tab:results-RoBERTa-base}
\end{table}

Table \ref{tab:results-RoBERTa-base} illustrates the results for the $RoBERTa_{Base}$ model. We observe an increase of $2.03\%$ in \textit{Exact Match} and $1.74\%$ in \textit{F1} when fine-tuning it with \textit{SQuAD} followed by \textit{Simple-SQuAD}, in contrast with the model when trained on only SQuAD. 

\begin{table}[h]
\centering
\resizebox{0.47\textwidth}{!}{%
\begin{tabular}{|l|l|r|r|}
\hline
\textbf{Model} & \textbf{Input} & \textbf{Exact} & \textbf{F1} \\ \hline
$RoBERTa_{Large}$ & SQuAD & 0.835 & 0.905 \\ \hline
$RoBERTa_{Large}$ & Simple-SQuAD & \textbf{0.852} & \textbf{0.917} \\ \hline
$RoBERTa_{Large}$ & Simple-SQUAD $\to$ SQuAD & 0.838 & 0.907 \\ \hline
$RoBERTa_{Large}$ & SQuAD $\to$ Simple-SQuAD & 0.836 & 0.908 \\ \hline
\end{tabular}%
}
\caption{Benchmarking Results for $RoBERTa_{Large}$.}
\label{tab:results-RoBERTa-large}
\end{table}

Table \ref{tab:results-RoBERTa-large} illustrates the results for the $RoBERTa_{Large}$ model. Similar to $RoBERTa_{Base}$, we observe an increase of $2.04\%$ in \textit{Exact Match} and $1.33\%$ in \textit{F1} when fine-tuning it with \textit{Simple-SQuAD} when compared to model when trained on \textit{SQuAD}. 

\begin{table*}[h]
\centering
\resizebox{\textwidth}{!}{%
\begin{tabular}{|l|l|l|}
\hline
{\textbf{Category}} &
  \textbf{Original} &
  \textbf{Transferred} \\ \hline
{\textbf{\begin{tabular}[c]{@{}l@{}}Inter-Event \\ Splitting\end{tabular}}} &
  \begin{tabular}[c]{@{}l@{}}Although his administrative abilities \underline{had been noticed},\\ on the eve of the U.S. entry into World War II he had\\ never held an active command above a battalion and\\ was far from \underline{being considered} by many as a potential\\ commander of major operations.\end{tabular} &
  \begin{tabular}[c]{@{}l@{}}Although his administrative abilities \underline{had been noticed}\\ on the eve of the U.S. entry into World War II he had\\ never held an active command above a battalion.\\ He was far from \underline{being considered} by many as a\\ potential commander of major operations.\end{tabular} \\ \hline
\textbf{\begin{tabular}[c]{@{}l@{}}Intra-Event \\ Splitting\end{tabular}} &
  \begin{tabular}[c]{@{}l@{}}Clark also \underline{claimed} that Abdul gave him preferential\\ treatment on the show due to their affair.\end{tabular} &
  \begin{tabular}[c]{@{}l@{}}Clark also \underline{claimed} that Abdul gave him preferential\\ treatment on the show. This was due to their affair.\end{tabular} \\ \hline
\end{tabular}%
}
\caption{Examples of event-based splitting.}
\label{tab:event-edit-analysis}
\end{table*}
\begin{figure*}[ht]
\begin{subfigure}{.49\textwidth}
  \centering
  \includegraphics[width=1\linewidth]{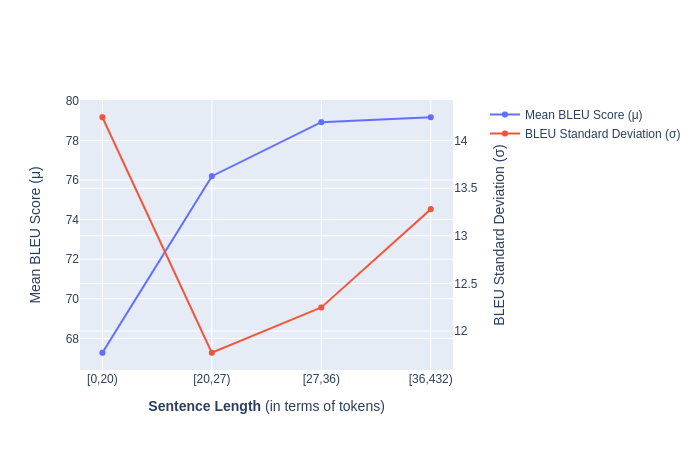}  
  \caption{BLEU}
  \label{fig:sub-first}
\end{subfigure}
\begin{subfigure}{.49\textwidth}
  \centering
  \includegraphics[width=1\linewidth]{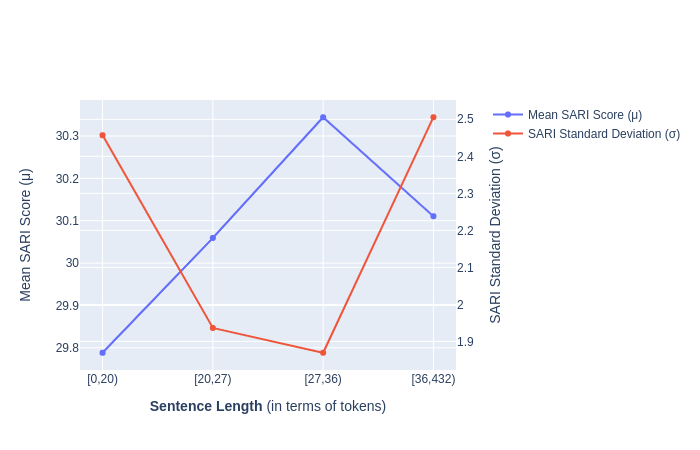}  
  \caption{SARI}
  \label{fig:sub-second}
\end{subfigure}
\newline
\begin{subfigure}{1\textwidth}
    \centering
    \includegraphics[width=.5\linewidth]{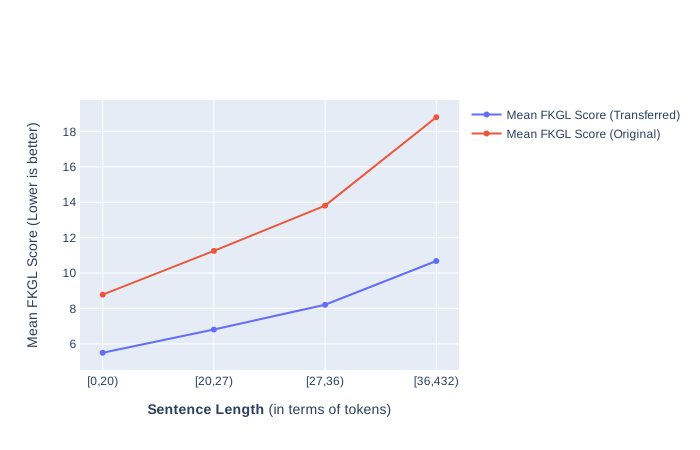}
    \caption{FKGL}
    \label{fig:sub-third}
\end{subfigure}
\caption{Plots illustrating sentence-length based transfer analysis}
\label{fig:sentence-length-distribution}
\end{figure*}
\section{Discussion}
\subsection{Edit Analysis}

We conduct an event-based analysis of the edits performed by the transfer model to convert the originally complex sentences into simpler forms. We use the definition of a linguistic event as defined in Pustejovsky~\cite{Pustejovsky1991}. We perform the analysis on $50$ parallel sentences in their pre and post simplification forms, classifying each edit into the following two classes: \textit{Inter-Event Splitting} and \textit{Intra-Event Splitting}. 

\textit{Inter-Event Splitting} denotes the type of edit in which the model splits two different events. As illustrated in Table~\ref{tab:event-edit-analysis}, the events of \textit{notice} and \textit{consider} are split into two different sentences, thus simplifying the complex sentence containing the two events. On the other hand, \textit{Intra-Event Splitting} is the type of edit in which all the simplified sentences contain the same event as the original sentence, as is illustrated for the event \textit{claim} in Table~\ref{tab:event-edit-analysis}. 

In our analysis, we found $32\%$ of the instances to show \textit{Inter-Event Splitting}, showing that our model can capture event boundaries. On the other hand, $60\%$ of the total instances show successful \textit{Intra-Event Splitting}, illustrating that the model can capture intra-event detailing boundaries. Interestingly, $8\%$ of the total instances displayed unsuccessful attempts of \textit{Intra-Event Splitting}, which can be improved in future work.

\subsection{Transfer Analysis}
For transfer analysis, we divide all the sentences in \textit{Simple-SQuAD} into four buckets based on the original sentence length split equally in terms of word-level tokenization ( $0-20$ tokens, $20-27$ tokens, $27-36$ tokens and $36-432$ tokens). We then compute the following three metrics on a sentence level: BLEU score, SARI scores, and FKGL scores. In Figure~\ref{fig:sentence-length-distribution}, we observe that the performance of the model varies with the sentence length. 

\begin{enumerate}
    \item \textbf{BLEU}: Sentence-preservation, measured through average BLEU score, is directly proportional to the sentence length. However, the standard deviation of the BLEU scores first decreases then increases as we increase the sentence length. 
    \item \textbf{SARI}: Lexical simplicity, measured through average SARI score, first increases then decreases as we increase the sentence length. However, the standard deviation of the SARI scores first decreases then increases as sentence length increases. 
    \item \textbf{FKGL}: Text readability, measured via sentence-level FKGL score, was computed for both transferred and original sentences. We observe that sentence-level FKGL score of transferred sentences is directly proportional to the sentence length. Whereas sentence-level FKGL scores for original sentences first decreases then increases with sentence length. Moreover, the mean sentence-level FKGL score for the transferred sentences was always lower than that for the original sentences regardless of the sentence length.
\end{enumerate}

\section{Conclusion}
In this work, we study the effect of text simplification in the comprehension based question-answering downstream task using the \textit{SQuAD} dataset.  For \textit{Simple-SQuAD} corpus creation, we use a transformers based style transfer model to transfer complex sentences to sequences of simple sentences while retaining the original meaning. We further use post-editing techniques to reduce noise from the dataset, followed by the use of heuristics to find required offsets for an answer in each question-answer pair. We prove the efficacy of our model using automated evaluation as well as human evaluation. We then benchmark \textit{Simple-SQuAD} using two different variants of RoBERTa and perform an ablation study to investigate the effects of text simplification using four different variations of input.  We prove that text simplification in the question-answering downstream task increases the predictive performance of the models. We further conduct edit-type analysis and sentence-length analysis to give insights about the transfer process. Future work may include improving style transfer performance using a more extensive corpus for text simplification and exploring effects of text simplification for other downstream tasks like text summarization, sentiment analysis.

\bibliography{anthology,custom}
\bibliographystyle{acl_natbib}

\end{document}